\documentclass[10pt,twocolumn,letterpaper]{article}

\usepackage{3dv}
\usepackage{times}
\usepackage{epsfig}
\usepackage{graphicx}
\usepackage{amsmath}
\usepackage{amssymb}
\usepackage{caption}
\usepackage{subcaption}
\usepackage{floatrow}
\usepackage{booktabs}

\newcommand{\otoprule}{\midrule[\heavyrulewidth]}
\usepackage[ruled]{algorithm2e}

\usepackage{color}
\usepackage{multirow}

\usepackage[pagebackref=true,breaklinks=true,letterpaper=true,colorlinks,bookmarks=false]{hyperref}

\threedvfinalcopy 

\def\threedvPaperID{156} 

\ifthreedvfinal\fi
\begin{document}

\title{Recalibration of Neural Networks for Point Cloud Analysis}

\author{Ignacio Sarasua, Sebastian  P{\"o}lsterl , Christian Wachinger \\
Artificial Intelligence in Medical Imaging (AI-Med)\\
Department of Child and Adolescent Psychiatry\\
Ludwig-Maximilians-Universit{\"a}t, Munich, Germany\\
{\tt\small ignacio@ai-med.de}
}

\maketitle

\begin{abstract}
Spatial and channel re-calibration have become powerful concepts in computer vision. Their ability to capture long-range dependencies is especially useful for those networks that extract local features, such as CNNs. While re-calibration has been widely studied for image analysis, it has not yet been used on shape representations. 
In this work, we introduce re-calibration modules on deep neural networks for 3D point clouds. We propose a set of re-calibration blocks that extend Squeeze and Excitation blocks~\cite{hu2018squeeze} and that can be added to any network for 3D point cloud analysis that builds a global descriptor by hierarchically combining features from multiple local neighborhoods. We run two sets of experiments to validate our approach. First, we demonstrate the benefit and versatility of our proposed modules by incorporating them into three state-of-the-art networks for 3D point cloud analysis: PointNet++\cite{qi2017pointnet++}, DGCNN~\cite{Wang2018a}, and RSCNN~\cite{Liu2019}. We evaluate each network on two tasks: object classification on ModelNet40, and object part segmentation on ShapeNet. Our results show an improvement of up to 1\% in accuracy for ModelNet40 compared to the baseline method. In the second set of experiments, we investigate the benefits of re-calibration blocks on Alzheimer's Disease (AD) diagnosis. Our results demonstrate that our proposed methods yield a  2\%  increase  in  accuracy  for  diagnosing  AD and a 2.3\% increase in concordance index for predicting AD onset with time-to-event analysis.
Concluding, re-calibration improves the accuracy of point cloud architectures, while only minimally increasing the number of parameters. 
\end{abstract}

\section{Introduction}
Deep neural networks are offering new possibilities for shape analysis as they can learn a shape representation that is optimal for the given task, instead of relying on pre-defined shape representations~\cite{Bronstein2017}. 
One of the most common networks for shape analysis is the PointNet~\cite{Qi2017}, which takes 3D point clouds as input. 
The core of the PointNet architecture are a multi-layer perceptron that processes all points, and a max pooling operation that creates a global signature of the whole point cloud. 
\begin{figure}
    \centering
    \includegraphics[width=\textwidth]{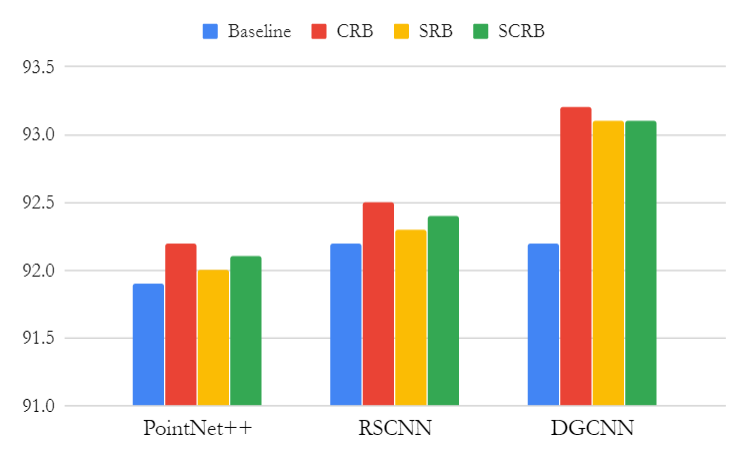}
    \caption{Classification accuracy (\%) on ModelNet40 with and without re-calibration. Blue: Baseline method without re-calibration blocks. Red: Baseline method re-calibrated using our channel-recalibration block (CRB). Yellow: Baseline method re-calibrated using our spatial-recalibration block(SRB). Green: Baseline method re-calibrated using our concurrent spatial-channel-recalibration block (SCRB)} 
    \label{fig:bars}
\end{figure}
By design, the PointNet architecture does not encode  local structures of the shape.
However, considering local structure and modeling spatial correlations has been at the core of the success of convolutional neural networks (CNNs). 
The multi-resolution hierarchy constructed by CNNs progressively captures features at increasingly larger receptive fields. 
This allows the network's filters to focus on finer details in the first layers and aggregate them in the later levels to generate more global features. 
Although the translation of these core concepts from CNNs to shape analysis is challenging, as  typical shape representations like point clouds and meshes do not possess an underlying Euclidean or grid-like structure, there have been multiple approaches proposed like PointNet++~\cite{qi2017pointnet++},DGCNN~\cite{Wang2018a}, and RS-CNN~\cite{Liu2019}.


An important addition to CNNs has been the re-calibration of feature maps, as introduced by the winner of the ImageNet 2017 challenge~\cite{hu2018squeeze}.
Re-calibration blocks explicitly model the interdependencies between the channels of the convolutional layers. 
As an extension, the re-calibration of spatial features has been proposed together with the concurrent  channel and spatial re-calibration~\cite{roy2018recalibrating}. 
It was demonstrated that re-calibration can significantly improve the overall performance with respect to the baseline architecture, while only marginally increasing the number of parameters.


Despite the wide success of re-calibration blocks in CNNs for image analysis, we are not aware of previous applications in shape analysis. 
In this work, we propose channel and spatial re-calibration blocks on point clouds and add them to a group of state-of-the-art point cloud architectures. 
Fig.~\ref{fig:bars} shows the results for object classification on ModelNet40 when  adding re-calibration blocks to PointNet++, RSCNN, and DGCNN. We can observe that re-calibration blocks increase the accuracy for all architectures. 
Next to object classification, we also evaluate the re-calibration blocks for object part segmentation where we also observe an improvement in performance with respect to the baseline methods. Finally, we study the effect of re-calibration on the diagnosis of Alzheimer's disease. First, we consider the classification of subjects with dementia and we further model the progression to dementia via progression analysis. 




 To summarize, the main contributions of this work are:
 \begin{enumerate}
     \item Integration of re-calibration blocks in state-of-the-art point cloud architectures.
     \item Introduction of a new spatial re-calibration block.
     \item Improvement in performance for all the baseline methods
     \item State-of-the-art results in Alzheimer's Disease diagnosis using 3D point cloud representations
 \end{enumerate}

\section{Related Work}
\textbf{Deep Learning for 3D Point Cloud Analysis.} One of the first deep learning methods for 3D point cloud analysis is PointNet~\cite{Qi2017}, which extracts descriptors for each point via a MLP, whose weights are shared across all points.
A subsequent max-pooling layer collapses individual descriptors into a global point cloud descriptor.
Due to this network architecture, PointNet is unable to capture fine-grained local geometric structures and is limited to solely describing the overall shape of an object.
To overcome this shortcoming, other methods built on top of this approach with the goal of capturing fine geometric structures \cite{qi2017pointnet++,PointWeb2019,yan2020pointasnl}.
Other approaches operate conceptually similar to traditional convolutional operators,
\cite{Hua2018,Wu2018,Xu2018,li2018pointcnn,Liu2019}  define convolutional kernels on a continuous space, where the neighboring points are weighted given the spatial distribution with respect to the center. In addition, another family of methods aims to capture local relations between the points by considering each point on the point cloud as a vertex of a graph  \cite{wang2019,rgcnn,Zhang2018,Liu2019DPAM,ClusterNet2019}. Capturing local structures within a point cloud, as done in conventional CNNs, is what allows the use of re-calibration blocks as the ones we propose in this work.

\textbf{Re-calibration Blocks for Image Analysis.} SENet~\cite{hu2018squeeze}, GENet~\cite{hu2018gather} and PSANet~\cite{zhao2018psanet} propose re-calibrating channel dependencies in a traditional 2D-CNN for image analysis. CBAM~\cite{woo2018cbam} also adds spatial re-calibration to the model. \cite{roy2018recalibrating} proposed concurrent spatial and channel re-calibration for medical image segmentation and \cite{rickmann2020recalibrating} extended these modules to work on 3D volumetric images. While re-calibration modules have been studied extensively in computer vision and medical image analysis, no previous work studied re-calibration for 3D shape analysis.

\textbf{Self-Attention on point clouds.} Self-attention networks have been recently applied to point cloud analysis in \cite{Yang2019,Liu2019Point2Sequencces}. However, unlike ours, these approaches propose an end-to-end attention network, rather than a set of blocks that can be added to other architectures in order to improve their performance. In addition, they do not consider channel re-calibration. 

\textbf{Anatomical shape analysis for Alzheimer's Disease Diagnosis.} Prior work in shape analysis for estimating discriminative models has mainly focused on the computation of handcrafted features~\cite{ng2014shape}, such as volume and thickness measures~\cite{costafreda2011automated}, medical descriptors~\cite{Gorczowski2007}, and spectral signatures~\cite{wachinger2016domain}. 
As an alternative, a variational auto-encoder was proposed to automatically extract features from 3D surfaces which were used for classification~\cite{Shakeri2016}.

\begin{figure}
    \centering
    \includegraphics[width=\textwidth]{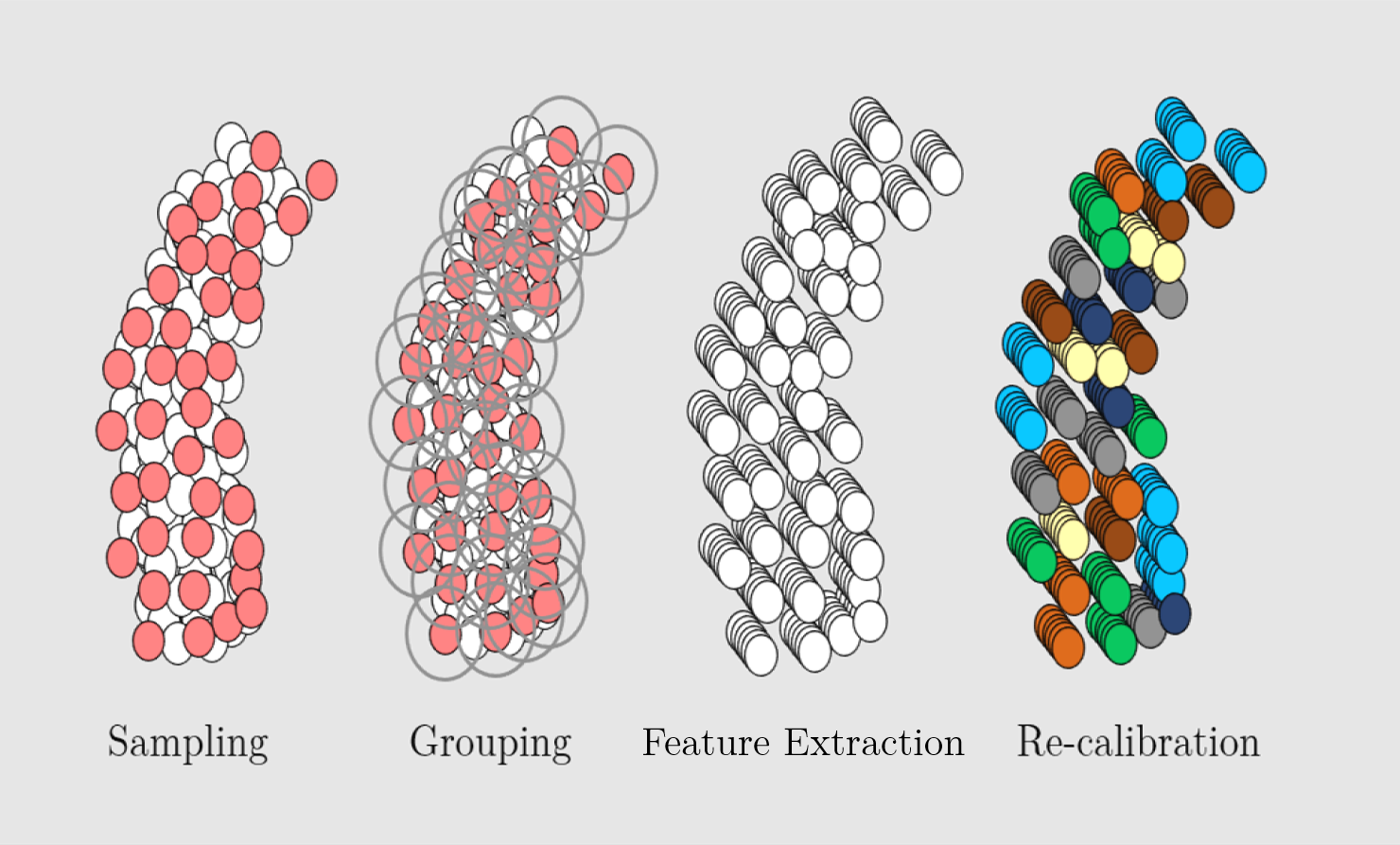}
    \caption{Main operations of a hierarchical network architecture. First a set of centroids are selected by \emph{sampling} a sub-set of points from the 3D point cloud. Then, in the \emph{grouping} a local neighborhood is defined. Finally a function $h$ \emph{extracts} local \emph{features} for each neighborhood. In addition, we propose \emph{re-calibrating} the features in order to include global information in each step.\label{fig:scheme}}
\end{figure}
\begin{figure*}
    \centering
    \includegraphics[width=0.9\textwidth]{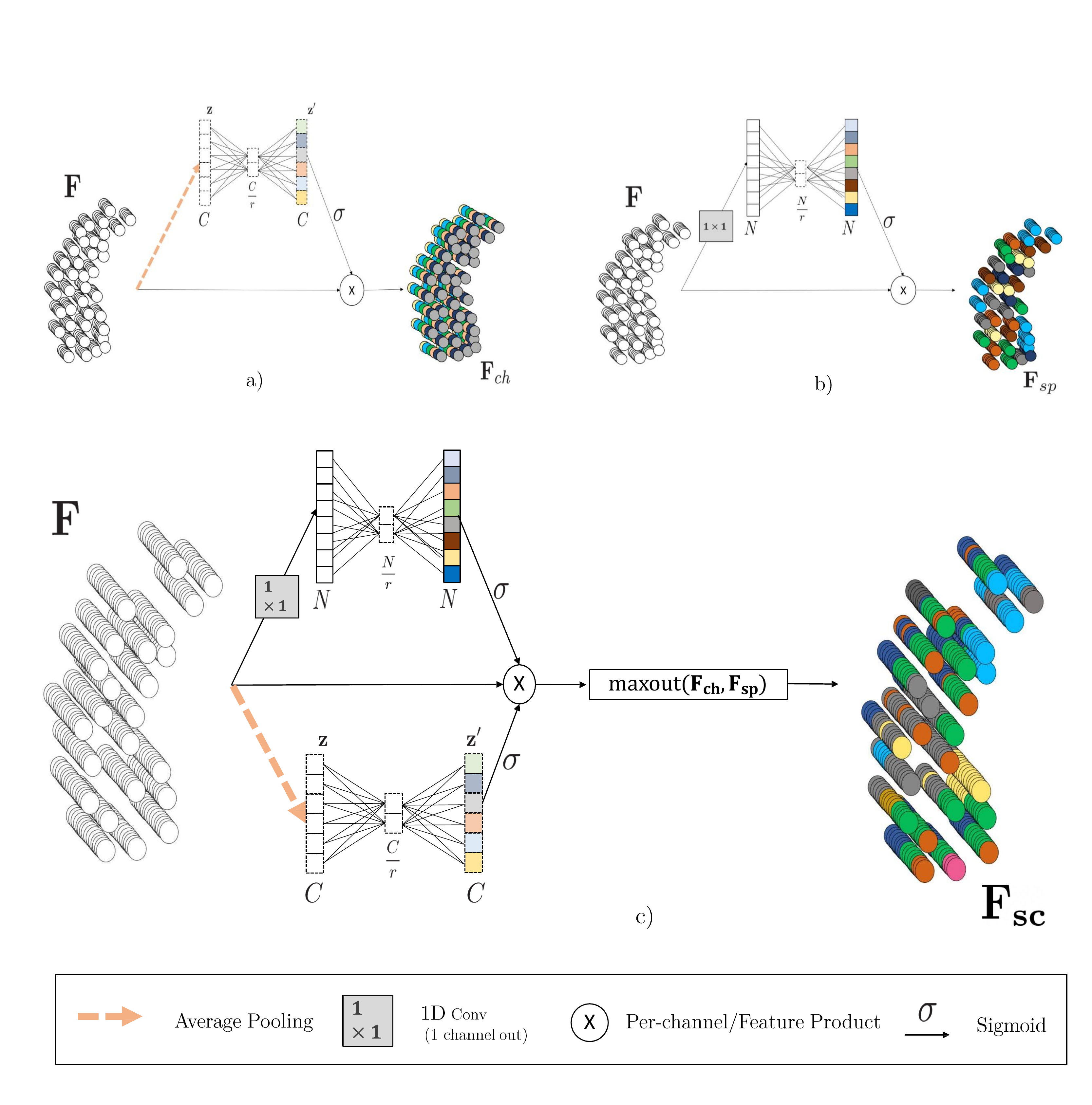}
    \caption{Overview of our re-calibration blocks. a) Channel Re-calibration Block (CRB) pools the channel information into a 1D vector $\mathbf{z}$ which is passed by a fully-connected bottle neck and sigmoid layer creating the re-calibration weights. The input channels are multiplied by the re-calibration weights,obtaining $\mathbf{F_{ch}}$. b) Spatial Re-calibration Block (SRB) yields the spatial information into a 1D vector $\mathbf{q}$, using a 1D convolutional layer with one channel as output. The vector is passed by a fully-connected bottle neck and sigmoid layer creating the spatial re-calibration weights. The input point descriptors are multiplied by the re-calibration weights, obtaining $\mathbf{F_{sp}}$. c) The combination of both, Spatial and Channel Re-calibration Block (SCRB),$\mathbf{F_{sc}}$, is formed by per element max-out operation of $\mathbf{F}_{ch}$ and $\mathbf{F}_{sp}$ .\label{fig:rec_blocks}}
\end{figure*}

\section{Methods}

While re-calibration blocks for traditional CNNs for image analysis are established, incorporating re-calibration blocks into networks for 3D point clouds is challenging, because 3D point clouds are sets, and not organized in a grid-like structure like images. Thus,
networks for 3D point cloud analysis require a very different architecture from a traditional CNN.

Given a 3D point cloud $\mathbf{P}= \{\mathbf{p}_1,...,\mathbf{p}_N\}$ with $\mathbf{p}_i=[x_i,y_i,z_i]$ being the coordinates of point~$i$, let $\mathbf{f}_i\in \mathbb{R}^C$ be a point descriptor associated with the $i^{\text{th}}$ point, $\mathbf{p}_i \in \mathbf{P}$, and $\mathbf{F}=[\mathbf{f}_1,...,\mathbf{f}_{N}]$, the matrix of all point cloud descriptors. A hierarchical point cloud architecture is usually composed by three main operations (Figure \ref{fig:scheme}): First, a subset of points are \emph{sampled} as centroids (e.g. Farthest Point Sampling). Once the centroids have been selected a local neighborhood is selected by \emph{grouping} the closest points to each centroid (e.g. query ball). Finally, the point descriptors associated to each neighborhood are obtained using a \emph{feature extraction} technique (e.g. convolution). While such networks are able to capture fine-grained geometric information, their local descriptors might be unaware of the global context.
We overcome this shortcoming by integrating global information into local descriptors via
re-calibration blocks.
In particular, we propose three different modules: channel re-calibration block (CRB), spatial re-calibration block (SRB) and the combination of both (SCRB), which are illustrated in Fig.~\ref{fig:rec_blocks}.

\subsection{Channel re-calibration (CRB).}\label{sec:c_rec}
Channel re-calibration blocks aim to encode global information into the point descriptors' channels. 
First, a global average pooling across $\mathbf{F} \in \mathbb{R}^{N \times C}$ is computed, yielding the vector $\mathbf{z} \in \mathbb{R}^{1 \times C}$ with its $j^{\text{th}}$ element being: 
\begin{equation}
    z_j = \frac{1}{N}\sum_i^N \mathbf{f_i}.
\end{equation}
The vector $\mathbf{z}$ is then transformed into $\mathbf{z}^\prime = \mathbf{W}_1^{\text{ch}}(\delta(\mathbf{W}_2^{\text{ch}}\mathbf{z}))$, with $\mathbf{W}_1^{\text{ch}} \in \mathbb{R}^{C \times \frac{C}{r}}$, $\mathbf{W}_2^{\text{ch}} \in \mathbb{R}^{\frac{C}{r}\times C}$ being the weights of two fully-connected layers and $\delta(.)$ the ReLU operator. $r$ is the reduction ratio of the bottleneck formed by the two fully connected layers.
$\mathbf{z}'$ is passed through a sigmoid layer, which creates re-calibration weights in the interval [0,1]. Consequently, $\sigma(z^\prime_j)$ can be seen as a relative measurement of the $j^{\text{th}}$-channel's importance with respect to the other channels and it can be used to obtain the channel re-calibrated  point cloud descriptors $\mathbf{F}_{\text{ch}} = [\mathbf{f'}_1,...,\mathbf{f'}_N]$, with $\mathbf{f'}_i =[\sigma(z'_1)f_{i,1},...,\sigma(z'_C)f_{i,C}], i\in \{1, \ldots, N\}.$


\subsection{Spatial re-calibration (SRB).}
While channel re-calibration blocks highlight the relevant channels, taking into account all the point descriptors at the same time, we would also like to study the opposite case: highlight the relevant areas of the shape taking into account all the channels. Hence, we need to map each local point descriptor $\mathbf{f}_i$ to a single value. This is achieved by passing $\mathbf{F}$ through a convolution operator with weights $\mathbf{W} \in \mathbb{R}^{1 \times C \times 1} $, which generates $\mathbf{q}=\mathbf{W}*\mathbf{F} \in \mathbb{R}^{N \times 1}$. $\mathbf{q}$ is transformed into   $\mathbf{q}' = \mathbf{W}_1^{\text{sp}}(\delta(W_2^{sp}\mathbf{q}))$, with $\mathbf{W}_1^{sp} \in \mathbb{R}^{N \times \frac{N}{r}}$, $\mathbf{W}_2^{\text{sp}} \in \mathbb{R}^{\frac{N}{r}\times N}$ being the weights of two fully-connected layers. The projection is now passed through a sigmoid layer to obtain the activations $\sigma(\mathbf{q}')$. Therefore, $\sigma(q^\prime_i)$ can be seen as the importance of the $i^{\text{th}}$ descriptor, associated with the centroid $\mathbf{p_i}$, with respect to the other descriptors. The spatially re-calibrated point cloud descriptors are formed as $\mathbf{F}_{\text{sp}}=[\sigma(q'_1)\mathbf{f_1},...,\sigma(q'_N)\mathbf{f_N}]$. Notice that we want to have the same property as for channel SE and thus use a bottle neck layer to force the network to only retain the most valuable information. This is in contrast to \cite{roy2018recalibrating}, which only used sigmoid, because a bottle-neck for 2D weight matrices is not straight-forward.



\subsection{Spatial and channel re-calibration (SCRB).}
As seen in~\cite{roy2018recalibrating}, max-out operators encourage a competitive behaviour between both blocks and yielded best results.  We therefore apply a per element max-out operator for combining both re-calibrated point cloud descriptors.
The simultaneously spatially and channel re-calibrated point cloud descriptors $\mathbf{F}_{\text{sc}}$ are: $\mathbf{F}_{\text{sc}} = \max(\mathbf{F}_{\text{sp}},\mathbf{F}_{\text{ch}})$.

\begin{figure}
    \centering
    \includegraphics[width=\linewidth]{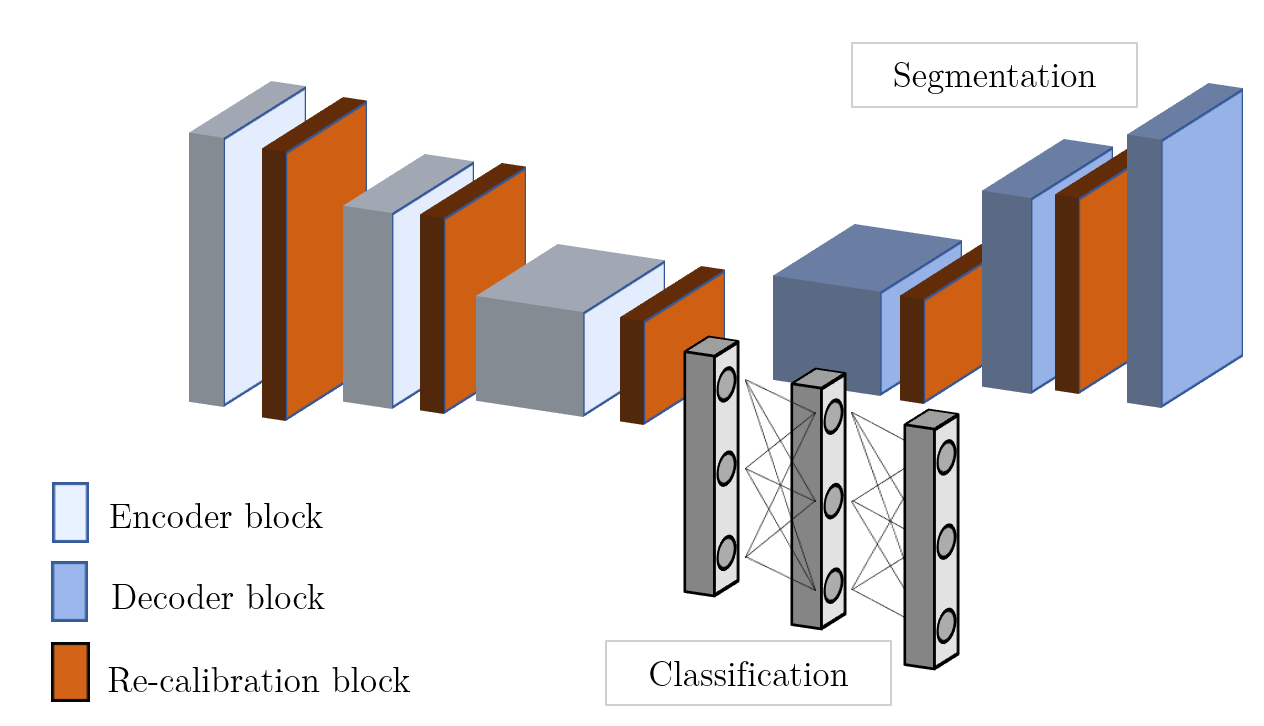}
    \caption{We position the re-calibration blocks after encoder and decoder layer in each network. The output of fully connected layers or pooling operations do not get re-calibrated, since they already include global information\label{fig:block_position}}
\end{figure}

\subsection{Baseline models}
Our proposed re-calibration modules are versatile and
can be incorporated in any network for 3D point cloud analysis that builds a global descriptor by hierarchically combining features from multiple local neighborhoods.
We evaluate our proposed blocks by extending three state-of-the-art networks: (i) PointNet++~\cite{qi2017pointnet++},
(ii) RSCNN \cite{Liu2019},
and (iii) DGCNN \cite{Wang2018a}.

\subsection{Implementation details}
All our blocks have been implemented in Pytorch. We fix the re-calibration factor,$r$, to $r=2$. As for the baseline methods, we incorporated the blocks in the author's Pytorch implementations.

\textbf{PointNet++} We trained every model for 200 epochs. We use Adam as the optimizer with a learning rate of 0.001, which decays by a factor of 0.7 every 20 epochs. The batch size is 24 samples.  

\textbf{DGCNN} We trained every model for 200 epochs. We use SGD as the optimizer with a learning rate of 0.1 using a Cosine Annealing scheduler and a momentum of 0.9. The batch size is 32 samples.  

\textbf{RS-CNN.} Every model is trained for 200 epochs. Adam is used as optimization algorithm. The learning rate begins with 0.001 and decays with a rate of 0.7 every 20 epochs. The momentum for BN starts with 0.9 and decays with a rate of 0.5 every 20 epochs. The batch size is 32.\footnote{While RS-CNN's code for their multi-scale grouping for segmentation is available, only the single-scale version for classification has been made publicly available so far.}

\section{Experiments}
We divide our experiments in two families of tasks: i) Object recognition (including classification and part segmentation) and ii) Alzheimer's Disease diagnosis (detection and prognosis). The former will evaluate the effect of the re-calibration blocks in different point-cloud neural networks and benchmarks, giving us a better understanding of their generalizability. The latter, given a fixed model, measures the effect of re-calibration on the study of Alzheimer's Disease progression, a field in which medical shape analysis is very relevant. 

\subsection{Object recognition}
Object recognition is a very relevant field within point cloud analysis, given its broad scope of applications (e.g. LIDAR systems). As such, a wide variety of benchmarks have been proposed to evaluate current methods. Two of the most popular ones are ModelNet40~\cite{Wu_2015_CVPR} for classification and ShapeNet~\cite{Yi2016} for part segmentation. For a fair assessment of the re-calibration blocks, we include them to three networks that capture local structures following three different approaches: i)by using a shared MLP (PointNet++) ii) by considering each point as a vertex of a graph (DGCNN) and iii) by emulating traditional convolutional operators (RS-CNN).  We evaluate the re-calibrated version of these three networks in the previously mentioned tasks.  The blocks are positioned after every encoder/decoder module excluding the fully-connected parts of the architectures
(see Figure~\ref{fig:block_position}).

\subsubsection{Object classification on ModelNet40}

\begin{table}
\caption{
\label{table:classification_modelnet40}%
Classification accuracy (\%) on ModelNet40 with and without re-calibration.}
\begin{tabular}{lc|ccc}
\toprule
Method            & Baseline         & +CRB         & +SRB  & +SCRB      \\
            \otoprule
PointNet++    & $91.9$          & $\mathbf{92.2}$          & $92.0$          & $92.1$          \\

DGCNN & $92.2$ & $\mathbf{93.2}$ & $93.1$ & $93.1$     \\
RS-CNN  & $92.2 $ & $ \mathbf{92.5}$  & $92.3$ & $92.4$     \\
	\bottomrule    
\end{tabular}
\end{table}

\textbf{Dataset} The dataset contains 12,311 meshed CAD models from 40 categories. 9,843 models are used for training and 2,468 models are for testing. 1024 points are sampled from each 3D shape and we follow the pre-processing and augmentation workflow described in each method's original paper in order to evaluate the effect of re-calibration alone.

\textbf{Results} Table \ref{table:classification_modelnet40} shows the performance of the baseline models together with each re-calibration block. We can observe a clear performance improvement for all three models. In particular, DGCNN benefits the most from the effect of re-calibration. We believe this is due to the fact that DGCNN builds the local neighborhood graph based on the distance between the points on the features space. Re-calibrating the feature space provides global information to the model helps it creating a more complete understanding of the shape. In addition, DGCNN concatenates all the intermediate features at the end before passing them through the MLP layers, adding even more weight to the re-calibration step. Also, the DGCNN network's design makes channel re-calibration more suitable, since while the number of channels remain small along the convolutional layers (64), the number of points is not reduced along the network. Consquently, the number of parameters for channel re-calibration marginally increases (around 5\%), while spatial re-calibration almost duplicates the number of parameters.  In opposition, PointNet++ and RS-CNN do not experience such an improvement. We believe this is due to the big correlation between the different local features.  First, PointNet++ uses Multi-scale grouping (MSG), which captures features on different resolutions and concatenates them in one feature vector. This has two effects on the re-calibration: i) The channel features are highly correlated since they describe the same region, biasing the channel re-calibration; ii) The size of the radius (up to 0.8) causes that each centroid shares most of its neighbors with the rest, which diminishes the effect of spatial re-calibration (similar to last layers of a CNN where the receptive field is very large, generating overlap between the regions). This last phenomenon is also observed in RS-CNN, even though we are comparing to its single scale version, it uses a fairly big radius for the query ball sampling (up to 0.32).  \label{sec:mn40_results}

\begin{table*}[t!]
\vskip 0.1in
\begin{center}
\resizebox{\columnwidth}{!}{
\begin{sc}
\begin{tabular}{l|c|ccccccccccccccccr}
\toprule
                & \textbf{mean} & areo & bag & cap & car & chair & ear & guitar & knife & lamp & laptop & motor & mug & pistol & rocket & skate & table  \\
                &          &          &.       &        &       &          & phone&       &          &          &           &            &         &           &           & board &          \\
\midrule
 \# shapes   &       &2690 &76  &55 & 898 & 3758 & 69 & 787 & 392 & 1547 & 451 & 202 & 184 & 283 & 66 & 152 & 5271 & \\
\midrule

PointNet++          & 81.8 & 82.4 & 79.0 & \textbf{87.7} & 77.3 & \textbf{90.8} & 71.8 & 91.0 & 85.9 & 83.7 & 95.3 & \textbf{71.6} & 94.1 & 81.3 & 58.7 & \textbf{76.4} & \textbf{82.6} \\
\midrule
Pointnet++ w/ CRB     & \textbf{82.1} & 82.7 & 79.0 & 86.4 & \textbf{78.6} & 90.5 & \textbf{76.9} & 91.0 & 85.9 & 83.3 & \textbf{95.7} & 69.6 & \textbf{94.9} & 80.8  & 60.4 & 76.0 & 82.4 \\
PointNet++ w/ SRB     & 81.6 & 82.5 & 75.6 & 83.0 & 78.5 & 90.1 & 76.1 & 90.8  & 85.8 & \textbf{84.2} & 95.6 & 69.8 & 94.7 & 80.4 & \textbf{60.5} & 76.2 & 81.8 \\
PointNet++ w/ SCRB    & 81.7 & \textbf{82.9} & \textbf{79.7} & 82.3 & 78.3 & 90.4 & 74.2 & \textbf{91.4} & \textbf{86.0} & 83.5 & 95.6 & 69.7 &\textbf{ 94.9} & \textbf{82.2} & 58.0 & 75.9 & 82.1 \\

\bottomrule
\end{tabular}
\end{sc}
}
\end{center}
\caption{Part segmentation results on ShapeNet part dataset for PointNet++ with the addition of re-calibration blocks (CRB, SRB, SCRB). Metric is mIoU(\%) on points.}
\label{table:part_pointnet}
\end{table*}

\begin{table*}[t!]
\vskip 0.1in
\begin{center}
\resizebox{\columnwidth}{!}{
\begin{sc}
\begin{tabular}{l|c|ccccccccccccccccr}
\toprule
                & \textbf{mean} & areo & bag & cap & car & chair & ear & guitar & knife & lamp & laptop & motor & mug & pistol & rocket & skate & table  \\
                &          &          &.       &        &       &          & phone&       &          &          &           &            &         &           &           & board &          \\
\midrule
 \# shapes   &       &2690 &76  &55 & 898 & 3758 & 69 & 787 & 392 & 1547 & 451 & 202 & 184 & 283 & 66 & 152 & 5271 & \\
\midrule
DGCNN                 &  80.3       & \textbf{83.7}     & 76.0           & 79.4             & 77.2              & 90.4              & 73.7              & \textbf{91.4}     & \textbf{88.5} & 84.2 & 95.8 & \textbf{59.1 }& 92.6 & 78.2 & 57.8 & 74.2 & \textbf{83.6}  \\
\midrule
DGCNN w/ CRB         & \textbf{80.5} & 83.1              & \textbf{79.7} & 80.4              & 76.6              & 90.2              & 75.4              & 91.0              & 87.6          & \textbf{84.9} & 95.5 & 57.4 & 92.8 & 80.4  & \textbf{59.0} & 72.8 & 81.6 \\
DGCNN w/ SRB             & 80.4      & 82.4              & 77.6          & 80.5              &\textbf{ 77.7}     & \textbf{90.5}     & 74.2              & 91.0              & 87.8          & 84.2 & \textbf{95.9} & 58.0 & 91.6 & \textbf{80.5} & 56.2 & \textbf{74.3} & 83.2 \\
DGCNN w/ SCRB             & 80.3     & 81.9              & 73.8          & \textbf{82.2}     & 77.2              & 90.0              & \textbf{75.7}     & 91.0              & 87.9          & 84.4 & 95.2 & 57.8 & \textbf{93.4} & 80.0 & \textbf{59.0} & 72.2 & 82.4 \\

\bottomrule
\end{tabular}
\end{sc}
}
\end{center}
\caption{Part segmentation results on ShapeNet part dataset for DGCNN with the addition of re-calibration blocks (CRB, SRB, SCRB). Metric is mIoU(\%) on points.}
\label{table:part_dgcnn}
\end{table*}

\begin{table*}[t!]
\vskip 0.1in
\begin{center}
\resizebox{\columnwidth}{!}{
\begin{sc}
\begin{tabular}{l|c|ccccccccccccccccr}
\toprule
                & \textbf{mean} & areo & bag & cap & car & chair & ear & guitar & knife & lamp & laptop & motor & mug & pistol & rocket & skate & table  \\
                &          &          &.       &        &       &          & phone&       &          &          &           &            &         &           &           & board &          \\
\midrule
 \# shapes   &       &2690 &76  &55 & 898 & 3758 & 69 & 787 & 392 & 1547 & 451 & 202 & 184 & 283 & 66 & 152 & 5271 & \\
\midrule
RS-CNN                 &  84.0       & 83.5     & 84.8           & 88.8             & \textbf{79.6 }             & 91.2              & \textbf{81.1 }             &  \textbf{91.6}    & \textbf{88.4} & \textbf{86.0} & \textbf{96.0} & 73.7& 94.1 & 83.4 & 60.5 & 77.7 & \textbf{83.6}  \\
\midrule
RS-CNN w/ CRB         & \textbf{84.2} & \textbf{84.4 } & 84.2 & 89.1              & 79.3              & 91.3              & 80.9              &\textbf{ 91.6}              & 87.6          & 85.3 & \textbf{96.0} & \textbf{75.4} & 94.2 & 82.4  & 62.8 & 77.1 & \textbf{83.6} \\
RS-CNN w/ SRB             & 84.0      & 83.8              & 85.8          & 87.2              &78.7     & 91.2     & 80.8  & 91.2              & 88.0          & 84.7 & 95.6 & 74.9 & 94.2 & 82.6 & \textbf{64.3} & 76.6 & \textbf{83.6 }\\
RS-CNN w/ SCRB             & \textbf{84.2 }    & 84.2              & \textbf{88.4 }         & \textbf{90.0 }    & 78.7              & \textbf{91.4 }             & 75.7     & 91.3           & 87.2          & 85.9 & 95.9 & 74.5 & \textbf{95.2} & \textbf{84.6} & 61.7 & \textbf{78.3} & \textbf{83.6} \\

\bottomrule
\end{tabular}
\end{sc}
}
\end{center}
\caption{Part segmentation results on ShapeNet part dataset for RS-CNN with the addition of re-calibration blocks (CRB, SRB, SCRB). Metric is mIoU(\%) on points.}
\label{table:part_rscnn}
\end{table*}
\subsubsection{Object part segmentation of ShapeNet}

\textbf{Dataset} The objective in this task is to assign each point from a point cloud set a part category label. The dataset contains 16,881 3D shapes. For PointNet++ and RS-CNN, we sampled 2048 points for a fair comparison with the results reported on their original work. Due to memory constraints, we had to reduce the number of points to 1024 for DGCNN. Since DGCNN does not reduce the number of points across the network, spatial re-calibration would be computationally too expensive for that originally proposed number of points (2048).

\textbf{Results} Tables \ref{table:part_pointnet},\ref{table:part_dgcnn},\ref{table:part_rscnn} summarize the results for PointNet++, DGCNN (with 1024 points) and RS-CNN, respectively. While the overall performance improvement is not as big as in classification, classes like earphone obtain an increase in performance up to 5\%. We also observe that Spatial re-calibration does not increase the performance of PointNet++ and RS-CNN. We believe this is caused by the high density of the point clouds (compared to the classification tasks) that leads to a more unstable training (given the high number of parameters of the MLP blocks).   


\begin{table*}[t!]
\vskip 0.1in
\begin{center}
\begin{sc}
\begin{tabular}{lcccc}

\toprule
Method            & Accuracy                        & Precision                       & Recall                        & F1-score                \\
            \otoprule
PointNet    & $0.780 \pm 0.025$          & $0.808 \pm 0.030$          & $0.766 \pm 0.041$          & $0.771 \pm 0.023$           \\
PointNet++*  & $0.826 \pm 0.028$          & $0.846 \pm 0.033$          & $0.821 \pm 0.041$          & $0.825 \pm 0.030$              \\
\midrule
PointNet++* w/ CRB  & $0.840 \pm 0.045$ & $\mathbf{0.849} \pm 0.042$ & $0.822 \pm 0.035$ & $\mathbf{0.829} \pm 0.037$    \\
PointNet++* w/ SRB  & $\mathbf{0.844} \pm 0.027$ & $0.827 \pm 0.047$          & $\mathbf{0.828} \pm 0.061$ & $0.826 \pm 0.030$   \\
PointNet++* w/ SCRB & $0.843 \pm 0.023$ & $0.830 \pm 0.042$          & $0.824 \pm 0.049$ & $0.808 \pm 0.029$       \\ 	
\bottomrule    
\end{tabular}
\end{sc}
\end{center}
\caption{HC-AD classification results for ADNI. *We use a modified version of PointNet++, where the number of centroids and radius of the query ball have been altered, as well as the grouping operation (using SSG instead of MSG).}
\label{table:adni_results}
\end{table*}

\begin{table}[t!]
\begin{center}
\begin{sc}
\begin{tabular}{lc}

\toprule
Method                           & c-index \\
            \otoprule
PointNet    & $0.677 \pm 0.028 $ \\
PointNet++*   & $0.692 \pm 0.046$    \\
\midrule
PointNet++* w/ CRB    & $ \mathbf{0.715} \pm 0.024$   \\
PointNet++* w/ SRB   & $0.702 \pm 0.031$   \\
PointNet++* w/ SCRB      & $0.701 \pm 0.038$   \\ 	\bottomrule    
\end{tabular}
\end{sc}
\end{center}
\caption{Results for progression analysis. *We use a modified version of PointNet++, where the number of centroids and radius of the query ball have been altered, as well as the grouping operation (using SSG instead of MSG).}
\label{table:adni_results_survival}
\end{table}

\subsection{Alzheimer's Disease Diagnosis}
Shape analysis plays and important tole in the study of neuro-degenerative diseases.
However, deep neural networks for 3D
point cloud analysis have not been as widely studied in this area as in computer vision. The current state-of -the-art for Alzheimer's disease prediction are based on PointNet \cite{gutierrez2018deep,polsterl2019wide}. Since PointNet's architecture does not allow for the integration of re-calibration blocks, we instead consider its hierarchical version, PointNet++. For Alzheimer's disease prediction, we modified the originally proposed architecture in the following ways. First, since changes due to Alzheimer's disease are small, we increase the number of centroids in the first and second layer to 1024 and 256 respectively.
Second, to reduce the overlap between neighborhoods of centroids and obtain locally distinctive features (as discussed in \ref{sec:mn40_results}), we reduce the radius of the query ball radius of the first and second layer to 0.1 and 0.2 respectively.
Third, we use Single Scale Grouping(SSG) instead of Multi-Scale Grouping(MSG), because we uniformly sample the points from the mesh of the anatomical structure to obtain a constant density of points across the structure.
We  evaluate  the  performance  of  the network  on  two  clinical tasks. In diagnosis, we use a classification model to distinguish healthy controls (HC) from patients diagnosed with Alzheimer's diseases. In progression analysis, we use a time-to-event model to predict the risk of progression from mild cognitive impairment (MCI) to AD. In our experiments on Alzheimer's Disease diagnosis, the channel re-calibration block was positioned after every encoder layer, while the spatial and spatial-channel re-calibration blocks are positioned only after the second encoder layer since their centroids are more sparse and ,therefore, there is less correlation between the local features.

\textbf{Dataset}
We used data from the Alzheimer's Disease Neuroimaging Initiative (ADNI)~\cite{Jack2008}.
For each patient, we segmented the magnetic resonance image of the brain using FreeSurfer~\cite{Fischl2012}.
We focused on the left hippocampus, given its  importance in AD pathology~\cite{thompson2004mapping}.
To obtain hippocampi point clouds, we 
sampled $N = 1,500$ points from the surface of the segmented structure.
All point clouds are centered on the origin and normalized so that their values fit in the unit sphere. We only use one baseline scan per patient (354 HC, 440 MCI and 275 AD) and validate our models using Monte Carlo cross-validation \cite{dubitzky2007fundamentals} with 10 sets. We split the data into training (70\%), validation (15\%), and test (15\%).

\begin{figure}
    \centering
    \includegraphics[width=\textwidth]{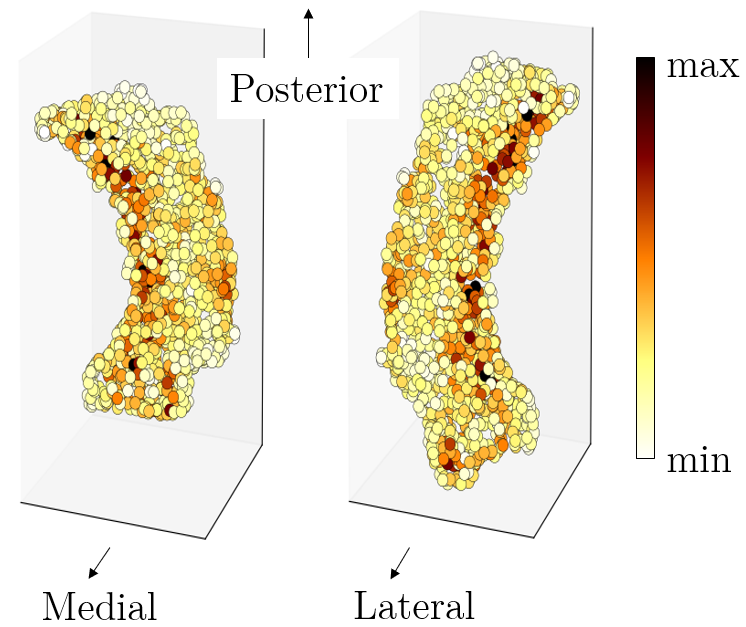}
    \caption{Main activations of the spatial re-calibration (SRB) for the HC-AD classification task. We  observe that the points on the medial part of the body in the subiculum area, the lateral part of the body in the CA1 area and the inferior part of the hippocampus head in the subiculum area are assigned higher weights, which is consistent with prior research on AD related changes \cite{lindberg2012shape}.}
    \label{fig:hippocampus-activations}
\end{figure}
\subsubsection{Alzheimer's Disease Diagnosis}
Table~\ref{table:adni_results} summarizes the predictive performance.
We observe that the PointNet++ without re-calibration outperforms PointNet by a 4.6\% in accuracy. Further, the integration of re-calibration blocks increases the accuracy of the prediction by an additional 1.8\% (among other classification metrics). In particular, spatial re-calibration has a much bigger effect compared to ModelNet40 (section \ref{sec:mn40_results}). We believe this is caused by the decrease in radius, which reduces the overlap between neighborhoods and allows capturing smaller changes, as observed in figure \ref{fig:hippocampus-activations}.
As a consequence, incorporating spatial, global information adds more value than it would for neighborhoods with large overlap.
Also, using SSG instead of MSG as grouping technique, make channel features become less redundant, and therefore CRB blocks have a much bigger effect on the final performance. 
Finally, the number of parameters only increases on a 5\%, 4\% and 9\% for CRB,SRB and SCRB respectively with respect to the baseline method. 


\subsection{Progression Analysis}

As a second experiment, we predict the progression from MCI to AD from censored and uncensored time-to-event data. We select $n=440$ ADNI MCI patients from whom we have follow-up data (136 uncensored and 334 censored) and make sure the ratio uncensored/censored is the same between training and testing for each of the cross-validation sets.  We define $t_i>0$  as the time to conversion for subject~$i$ and $c_i>0$ as censoring time ($c_i=\infty$ for uncensored data). Since we only have right-censored data (baseline scans are available for all subjects), we define $y_i = \min(t_i,c_i)$ and $\delta_i=I(t_i\leq c_i)$ for every patient, with $I(\cdot)$ the indicator function. We modify our model to predict a risk score instead of a classification label. As loss function, we use a deep neural network extension of Cox's proportional hazards model \cite{faraggi1995neural}
\begin{equation}
    \arg \min_{\Theta} \sum_{i=1}^n \delta_i \left[h(\mathbf{P}_i|\Theta) - \log \left(\sum_{j\in \mathcal{R}_i}\exp(h(\mathbf{P}_j|\Theta)) \right) \right]
\end{equation}
with $\Theta$ being the parameters of the network, $h$, and  $\mathcal{R}_i=\{j|y_j \geq t_i\}$ the risk set, i.e., the subset of patients who were not diagnosed with AD nearing $t_i$.
To evaluate the different models, we use the \emph{c-index} \cite{harrell1996multivariable}. Table \ref{table:adni_results_survival} summarizes the performance of each method across all the sets. These results echo our previous results in the classification task: incorporating re-calibration blocks leads to a significant improvement compared to the baseline methods.

\section{Conclusions}
In this work, we have proposed versatile channel and spatial re-calibration blocks for deep neural networks for 3D point cloud analysis.
We showed that our blocks can be integrated into three state-of-the-art networks: PointNet++, DGCNN, and RS-CNN.
We evaluated our proposed \emph{re-calibration} blocks on two computer vision benchmarks for object recognition, and on a medical dataset on Alzheimer's disease.
Our results show a significant improvement for object classification on ModelNet40 and object part segmentation on ShapeNet. In addition, our proposed methods achieve state-of-the-art performance on Alzheimer's disease diagnosis with 3D point clouds, while marginally increasing the number of parameters of the existing methods. 
In addition, we have visualized the re-calibration maps for a better understanding of the functionality of these methods.

{\small
\bibliographystyle{ieee}
\bibliography{egbib}
}

\end{document}